\documentclass[letterpaper, 10 pt, conference]{ieeeconf}
\IEEEoverridecommandlockouts                

\usepackage[noformatting,nogeometry]{stdcommands}
\usepackage[]{footmisc}
\usepackage{graphicx}
\usepackage{dsfont} 
\usepackage{amsthm}
\usepackage{amsmath}
\usepackage{amssymb}
\usepackage{url}
\usepackage[utf8]{inputenc} 
\usepackage[format=hang,singlelinecheck=false,caption=false,font=footnotesize]{subfig}
\usepackage[ruled,linesnumbered]{algorithm2e}
\usepackage{multirow} 
\usepackage{scalerel}
\usepackage[noend]{algpseudocode}
\usepackage[english]{babel}
\usepackage{mathrsfs}
\usepackage{bm}
\usepackage{hhline}
\usepackage{placeins}

\title{DRF: A  Framework for High-Accuracy Autonomous
Driving Vehicle Modeling}
\author{%
    \parbox{\linewidth}{\centering
        Shu Jiang$^1$,
        Yu Wang$^1$,
        Longtao Lin,
        Weiman Lin,
        Yu Cao,
        Jinghao Miao,
        Qi Luo$^2$
    }%
    \thanks{$^1$ Equally contributed to this paper}
    \thanks{$^2$ Corresponding author}
    \thanks{All Authors are with Baidu USA LLC,
        1195 Bordeaux Dr, Sunnyvale, CA 94089
            {\tt\small luoqi06@baidu.com}}%
}

\begin{document}
\maketitle
\thispagestyle{empty}
\pagestyle{empty}

\begin{abstract}

    An accurate vehicle dynamic model is the key to bridge the gap between simulation and real road test in autonomous driving. In this paper, we present a Dynamic model-Residual correction model Framework (DRF) for vehicle dynamic modeling. On top of any existing open-loop dynamic model, this framework builds a Residual Correction Model (RCM) by integrating deep Neural Networks (NN) with Sparse Variational Gaussian Process (SVGP) model. RCM takes a sequence of vehicle control commands and dynamic status for a certain time duration as modeling inputs, extracts underlying context from this sequence with deep encoder networks, and predicts open-loop dynamic model prediction errors. Five vehicle dynamic models are derived from DRF via encoder variation. Our contribution is consolidated by experiments on evaluation of absolute trajectory error and similarity between DRF outputs and the ground truth. Compared to classic rule-based and learning-based vehicle dynamic models, DRF accomplishes as high as $74.12\%$ to $85.02\%$ of absolute trajectory error drop among all DRF variations.
\end{abstract}

\IEEEpeerreviewmaketitle

\section{Introduction}

Autonomous driving has been emerging as a primary impetus and \emph{test bed} of novel artificial intelligence technologies. As one of the most influential open source autonomous driving platforms, \emph{Baidu Apollo} builds a professional self-driving simulator \emph{Apollo Dreamland} to execute  \emph{control-in-the-loop} simulation for fast iteration of planning and control algorithm. The control-in-the-loop simulator emulates dynamic behaviors of the ego vehicle and its interactions with other vehicles/pedestrians in thousands of high-fidelity scenarios. Consequently, an accurate vehicle dynamic model is crucial to address the well-known \emph{Sim-to-Real} transfer problem in robotics and control community.



Traditionally, the analytical modeling (or, rule-based modeling) methods make use of either dynamic equations (for parametric models) or frequency-domain dynamic responses (for non-parametric models) to identify the vehicle dynamics, which suffer a series of technical barriers, including: 1)~\emph{accuracy limitation}: the limited model order usually cannot fully cover the complex vehicle dynamics and also, the highly-nonlinear, multivariate vehicle model raises the difficulty of system identification and further degrades the modeling accuracy; 2)~\emph{lack of accuracy evaluation}: the deficiency of metrics and quantitative error analysis usually makes it difficult to evaluate how accurate/confident the model outputs are; and hence, the \emph{Sim-to-Real} transferring performance cannot be forecasted. In the autonomous driving field, the large-scale deployment of the dynamic modeling for a number of vehicles (with different vehicle models, mechanical structures, etc.) significantly amplifies these technical issues.

\subsection{Related Work}
\label{subsection:related_work}
Early researches on dynamic modeling focused on analytically addressing the system identification (ID) on linear systems~\cite{gevers2006linearID} and then extended to nonlinear systems ~\cite{gordon1993novel}~\cite{nelles2013nonlinear}~\cite{rong2006sequential}, from the viewpoint of control theory and combined with machine learning (ML) technology ~\cite {ghahramani1999learning}. Due to limited prior knowledge on system physics and limited dimensions of the state-space systems, it was difficult for these classic system ID methods to completely capture the overall dynamics of the complex mechanical-electrical systems. Currently, the most widely used methodologies for fitting nonlinear dynamic systems are Gaussian process (GP) and neural networks (NN)~\cite{Agudelo2020dataset}. The GP-based methods (e.g., PILCO method which presented the data-efficient reinforcement learning algorithms~\cite{Deisenroth201pilco}), were employed to well solve the dynamics-learning problems~\cite{romeres2019semiparam}; however, the computational complexity placed restrictions on their applications in large data sets. To overcome the computational issue, Sparse Variational GP (SVGP) method had been developed to incorporate stochastic variations inference to enable mini-batch learning~\cite{Agudelo2020dataset}~\cite{hensman2015scalable} and further, to correct the sensor measurement errors in dynamic modeling~\cite{brossard2019learning}. On the other hand, Deep NN were introduced to address dynamic modeling and parameter ID~\cite{su2020deepnn}~\cite{zhang2020sufficient}. The successful application of SVGP and NN methods inspired us to integrate them for a more accurate algorithm in this paper. 

Some recent research work had brought NN or ML methods into the vehicle dynamic modeling. The well-designed two-hidden-layer NN integrated with advanced controls, were applied to identify the one-fifth scale rally car dynamic models~\cite{Williams2017MPC} and the helicopter dynamic models~\cite{Punjani2015helicopter}. A Gaussian Process Regression (GPR) model was trained to predict the driver's actions and vehicle states on a certain segmented track in racing car games~\cite{georgiou2015predicting}. These model architectures were relatively shallow and thus, had difficulty to accurately capture the vehicle dynamics for autonomous driving applications. In some more recent work~\cite{Devineau2018vehicle}, Deep NN involving Convolutional NN (CNN) and Multi-Layer Perceptron (MLP) were trained to compute vehicle controls; however, the classic state-space vehicle models were still employed and hence, the potential improvement the deep learning may bring in was limited. Further, our \emph{Baidu Apollo} team has proposed a learning-based dynamic modeling method~\cite{apollo_2019}, in which the NN methods were leveraged to model the complicated vehicle dynamics. It has higher accuracy and lower cost compared to the rule-based model and is suitable for large-scale autonomous driving system deployment. However, the accuracy of this dynamic model is still limited by its open-loop structure.




\subsection{Contributions}
\label{subsection:main_contributions}

To improve the dynamic modeling accuracy, we propose a new modeling framework, named D (dynamic model) R (residual correction model) F (framework). DRF combines two models: 1) an open-loop dynamic model, which can be either an analytic dynamic model or an existing learning-based one (as introduced in~\cite{apollo_2019}), to predict the baseline position outputs; 2) a residual correction model (RCM), which integrates a deep encoder of different varieties with a SVGP model, to correct the prediction error from the open-loop dynamic model.


Following the basic RCM structure, five different DRF models are developed by employing different deep encoders. We compare the mean absolute trajectory errors (m-ATE) from the ground truth, between the trajectories generated by DRF-derived models and the one generated by open-loop dynamic model. We show that our best performed DRF model achieves up to $74.12\%$, $76.65\%$ and $85.02\%$ drop in m-ATE at $1s$, $10s$, and end of trajectories ($\sim60s$) under various scenarios compared with open-loop dynamic model. In addition, the DRF prediction error bound information are provided to support vehicle control design.

Besides ATE, we introduce five complementary metrics to evaluate the performance of dynamic model in autonomous driving, which jointly form a set of evaluation standards for gauging the model accuracy. Our models achieve significant performance increase under these standards as well.

\section{Vehicle Dynamic Modeling Framework}

\begin{figure}[!ht]
    \includegraphics[width=0.48\textwidth]{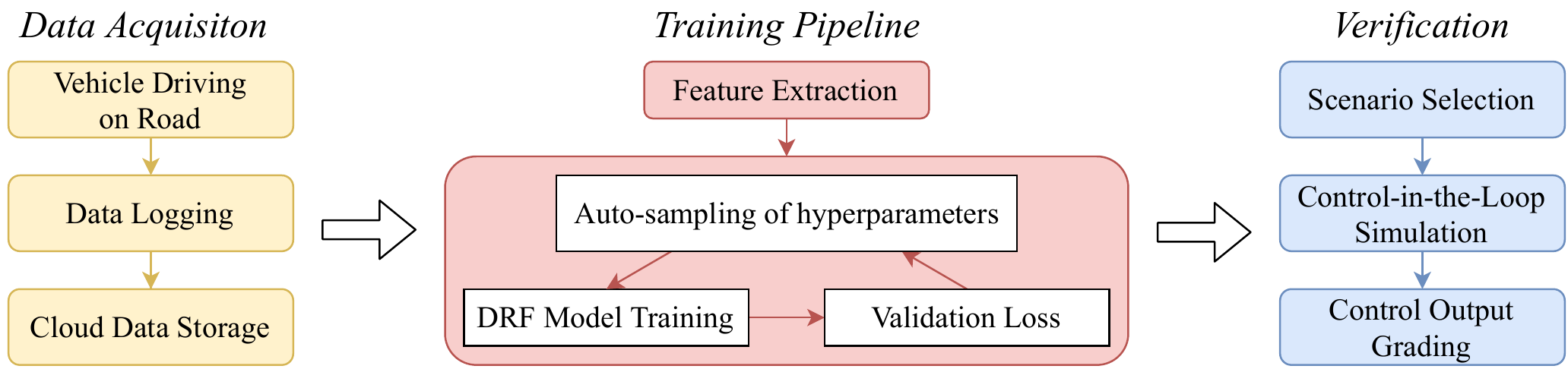}
    \caption{Apollo DRF Machine Learning Platform}
    \label{fig:DM20_architecture}
\end{figure}

DRF is trained and evaluated at Apollo DRF machine learning platform.
This platform (Fig.~\ref{fig:DM20_architecture}) can be broken down into three parts: \emph{1) data acquisition}, which provides data collection, logging and storage services; \emph{2) training pipeline}, which includes large-scale data processing, feature extraction, model training, hyperparameter-tuning and offline evaluation; \emph{3) verification}, which contains the control-in-the-loop simulation and performance grading.

\subsection{Data Acquisition}
There are two types of data to collect: one for dynamic model training and validation; one for model performance evaluation on continuous trajectories.

\paragraph{Data for Training and Validation}
Either manual driving data or autonomous driving data (as long as the throttle/brake/steering behaves the same under control commands) can be taken as training data. In order to make training data evenly distributed in input (control command) and output (vehicle state) spaces, the same feature collection and extraction standard is applied as described in our previous work~\cite{apollo_2019}.

\begin{figure}[!h] 
    \centering
    \includegraphics[width=0.48\textwidth]{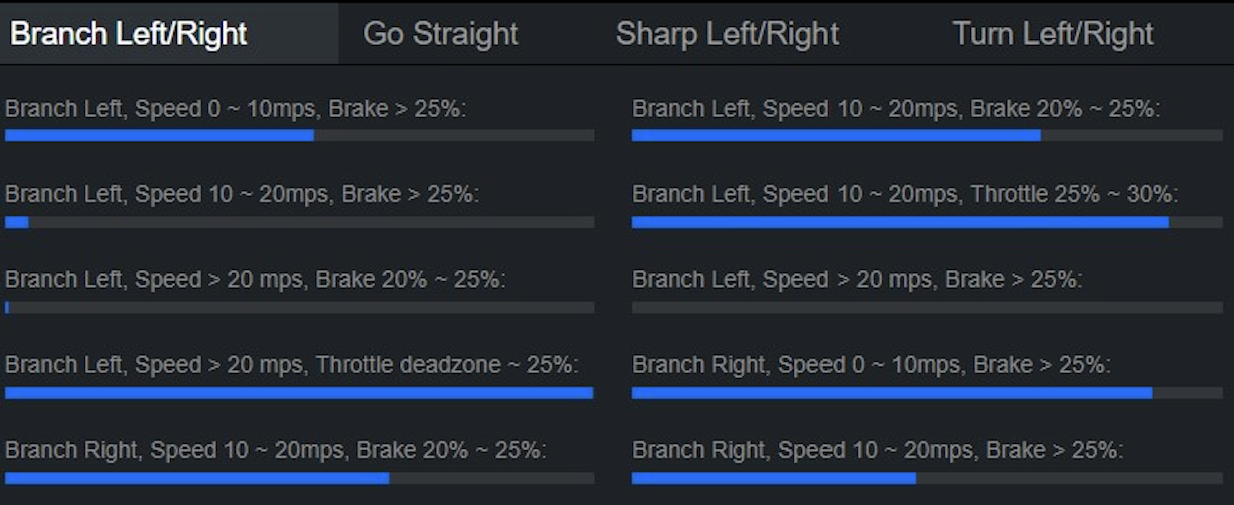}
    \caption{Apollo Dynamic Model Data Collection Interface}
    \label{fig:collection_finish.png}
\end{figure}

Fig. \ref{fig:collection_finish.png} shows a portion of Apollo's training data collection GUI. Each blue bar indicates the collection progress of a specific category.  Based on the control commands and vehicle states, we have a total of 102 categories.

\paragraph{Data for Trajectory Evaluation}
Besides validation datasets, a set of eight driving scenarios, named golden evaluation set, is picked to evaluate model performance on typical driving maneuvers.  Golden evaluation set includes left/right turns, stop/non-stop and zig-zag trajectories, the combination of which covers normal driving behaviors. We also prepare a 10-minutes long driving scenario to check model performance on a normal urban road with start/stop at traffic light, left turn, right turn, change lane and different speed variations etc. Unlike training and validation datasets, evaluation datasets are continuous time sequential datasets.

\subsection{DRF Training Pipeline}

\begin{figure}[h]
    \centering
    \includegraphics[width=0.48\textwidth]{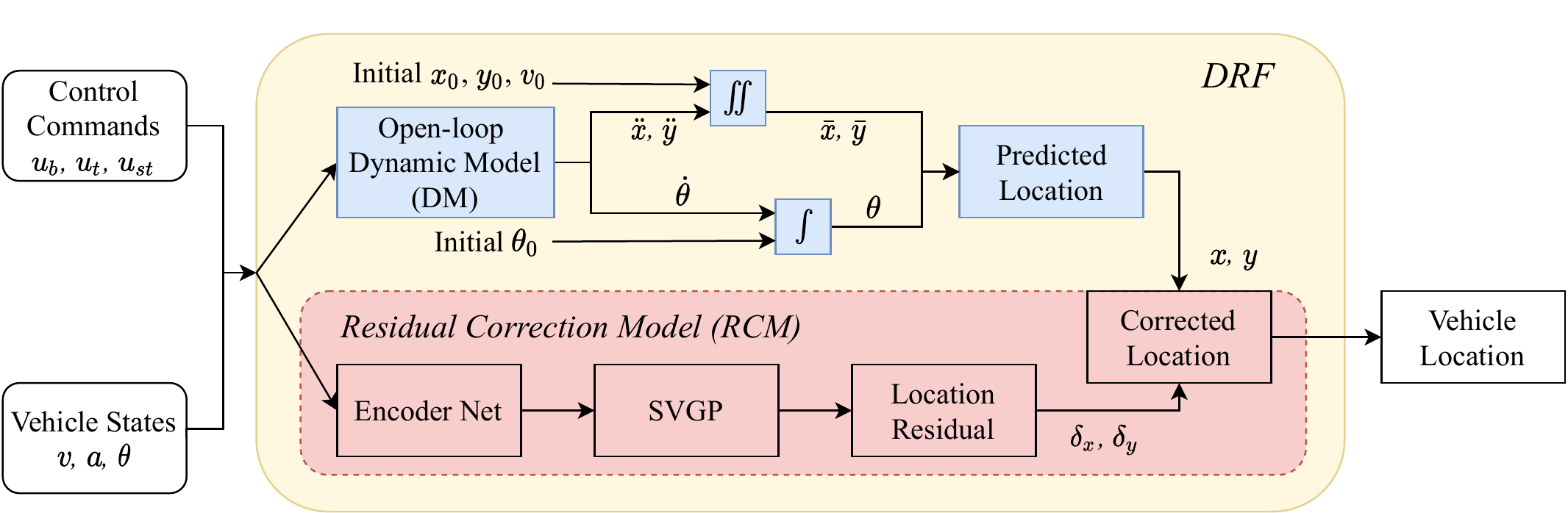}
    \caption{Dynamic residual correction Framework (DRF) Structure}
    \label{fig:DM20_structure}
\end{figure}

Components of DRF, as shown in Fig.~\ref{fig:DM20_structure}, are 1) Dynamic model (DM) 2) Residual Correction model (RCM). 

\subsubsection{Dynamic Model (DM)}
DM models the open-loop vehicle dynamics system.  Both rule-based (DM-RB) and learning-based (DM-LB) models can be integrated to DRF. Here we choose a pre-trained MLP model \cite{apollo_2019}, which passes a $5$-dimensional input to  an $8$-dimensional fully-connected layer with ReLU activation and produces a $2$-dimensional output. As shown in Fig.~\ref{fig:DM20_structure}, outputs from DM are integrated over time to produce the vehicle's location in simulation.



\subsubsection{Residual Correction Model (RCM)}
RCM corrects the error between vehicle location from DM and that from ground truth utilizing previous $N$ time step data.
At $t=i+N$, inputs of RCM consist of control commands and vehicle dynamic states over previous $N$ steps, denoted as $\bm{u} = \left[\bm{u}_{i+1}, \ldots, \bm{u}_{i+N}\right]$, where $\bm{u}_{i}=\left[u_{t,i}, u_{b,i}, u_{st,i}, v_{i}, a_{i}, \theta_{i} \right]^{T}$ at $t=i$. The inputs $u_{t}$, $u_{b}$ and $u_{st}$ represent throttle, brake and steering control commands respectively. $v$, $a$ and $\theta$ are the vehicle speed, acceleration and heading angle. Outputs of RCM are location residuals at $t = i+N$, denoted as $\bm{\delta} = \left[\delta_{x, i+N}, \delta_{y, i+N} \right]^{T}$.  The corrected vehicle location at $t=i+N$ is 
 $x_{i+N} = \bar{x}_{i+N}+\delta{x}_{i+N}$ and $y_{i+N} = \bar{y}_{i+N}+\delta{y}_{i+N}$, where $\pb{\bar{x}_{i+N},\bar{y}_{i+N}}$ is location integrated from DM.

\begin{figure}[h]
    \includegraphics[width=0.48\textwidth]{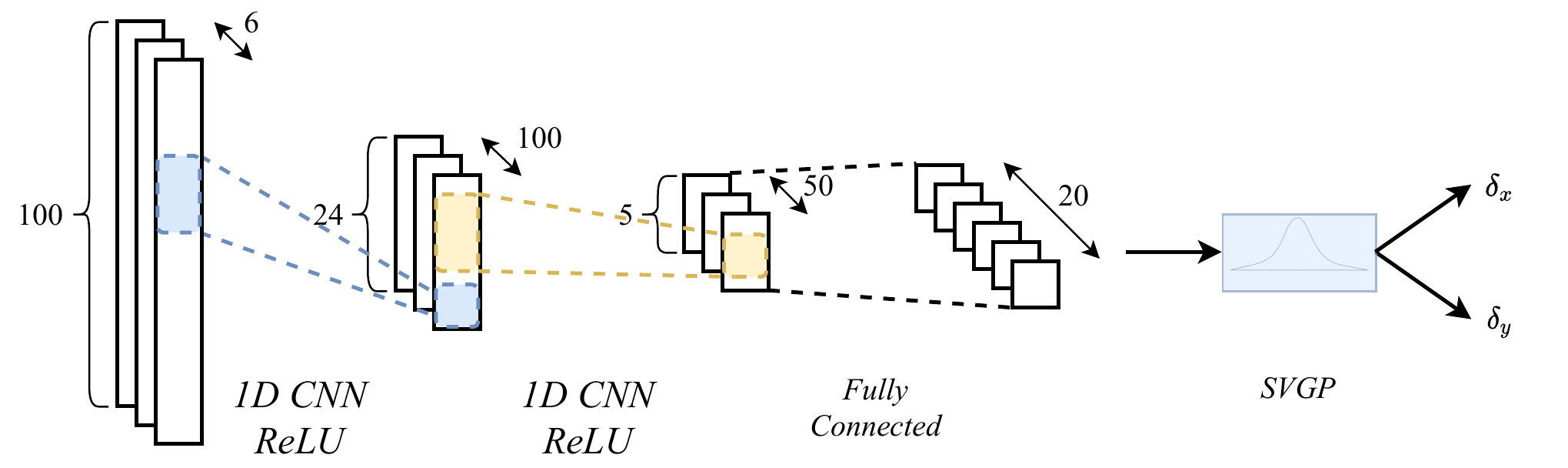}
    \caption{Residual correction module structure with CNN encoder}
    \label{fig:CNN}
\end{figure}

\begin{figure}[h]
    \includegraphics[width=0.48\textwidth]{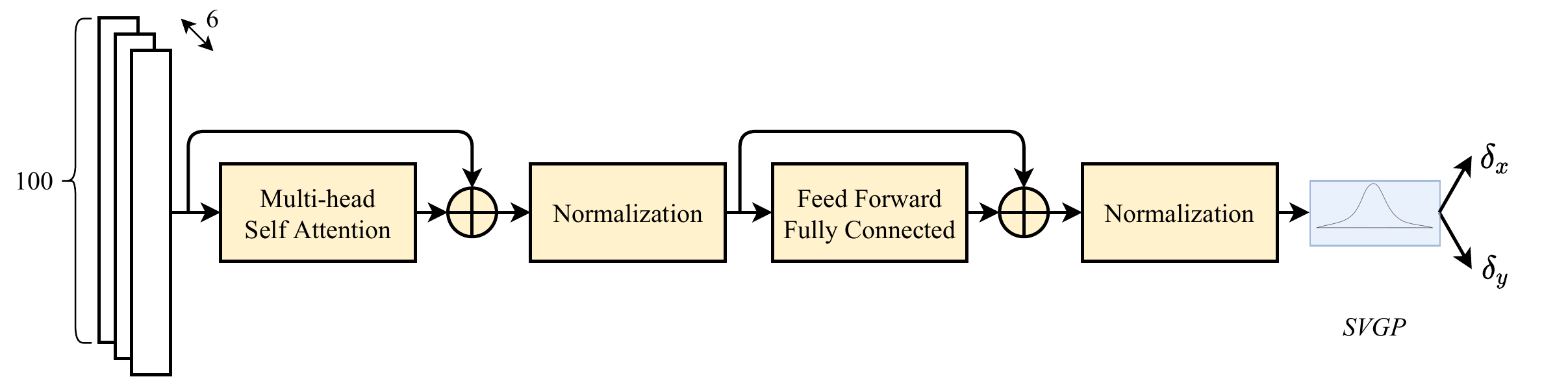}
    \caption{Residual correction module structure with transformer encoder}
    \label{fig:transformer}
\end{figure}

\paragraph{Encoder Variance}
The encoder network reduces the dimensions of RCM input sequence $\bm{u}_1, \ldots, \bm{u}_n$ by mapping them to a latent space.  Different encoder networks affect the accuracy of the final model.  
We select five different encoder networks, which are CNN, Dilated CNN \cite{chang2017dilated}, stand-alone attention \cite{parmar2019stand}, LSTM \cite{sun20183dof} and transformer encoder \cite{vaswani2017attention}. Their structure and hyperparameters are shown in Table~\ref{table:encoder_params}.
Fig.~\ref{fig:CNN} -~\ref{fig:transformer} show RCM model structures with CNN encoder and transformer encoder respectively.

\begin{table}[!ht]
    \caption{Encoder Network Structures and Hyperparameters}
    \centering
    \begin{tabular}{p{1.2cm}  p{2.8 cm}  p{2.2cm} p{0.5cm}}
        \hline
        Encoder                       & Structure              & Hyperparameters   & Value
        \\\hhline{=:===}
        \multirow{3}{*}{CNN}          & 2 * 1-D Conv           & Kernel size       & 6
        \\ \cline{3-4}
                                      & with ReLU              & Stride size       & 4
        \\\cline{2-4}
                                      & 1 FC                   & Filter No.        & 250
        \\\hline
        \multirow{7}{*}{Dilated}      & 1 * $1$-D Dilated Conv & Kernel size       & 6
        \\\cline{3-4}
        \multirow{7}{*}{CNN}          & with ReLU              & Dilation size     & 5
        \\\cline{3-4}
                                      &                        & Stride            & 4
        \\\cline{2-4}
                                      & 1 * $1$-D Dilated Conv & Kernel size       & 6
        \\\cline{3-4}
                                      & with ReLU              & Dilation size     & 1
        \\\cline{3-4}
                                      &                        & Stride            & 4
        \\\cline{2-4}
                                      & 1 FC                   & Filter No.        & 200
        \\\hline
        \multirow{2}{*}{LSTM}         & 1 LSTM layer           & Hidden layer size & 128
        \\\cline{2-4}
                                      & 1 FC                   & Filter No.        & 128
        \\\hline
        \multirow{2}{*}{Attention}    & 2 * Dot Attention      & Kernel size       & 5
        \\\cline{3-4}
                                      &                        & Stride size       & 5
        \\\cline{2-4}
                                      & 1 FC                   & Filter No.        & 200
        \\\hline
        \multirow{2}{*}{Transformer } & 1 Transformer Encoder  & Head dimension    & 1
        \\ \cline{3-4}
                                      &                        & Feed Forward Dim   & 1024
        \\\hline
    \end{tabular}
    \label{table:encoder_params}
\end{table}

\paragraph{SVGP} The encoded inputs, denoted as $\bm{z}$, are fed to SGVP to generate a final prediction of location residual.
The relation between the input sequence $\bm{z}$ and the residual $\bm{\delta}$ is assumed to satisfy a multivariate Gaussian process.
\begin{equation}
    \begin{aligned}
        \delta(\bm{z}) = \mathcal{GP}\pb{m(\bm{z}\pb{\bm{u}_i}), k(
        \bm{z}\pb{\bm{u}_i},\bm{z}\pb{\bm{u}_i})},
    \end{aligned}
\end{equation}
where $\bm{z}$ is the latent space formed by encoded inputs $\bm{u}$, $m(\bm{z}) = \mathbb{E}\sqparen{f(\bm{z})}$ and  $ k\pb{\bm{z},\bm{z}} = Cov\pb{f\pb{\bm{z}},f\pb{\bm{z}}}$ represent mean and covariance respectively. 


The offline training aims to find the GP model which maximizes the log likelihood $\log{p\pb{\bm{\delta}|\bm{z}}}$.  Considering the data dimensions are relatively large, we choose SVGP method and implement with GPyTorch \cite{gardner2018gpytorch}. SVGP takes $l$ inducing points, where $l$ is less than batch size, denoted as $\bm{U}$, and approximates the actual distribution of output $f\pb{\bm{u}}$ as
\begin{equation}
    \begin{aligned}
        q\pb{f(\bm{U})} = \int{p\pb{f\pb{\bm{U}}| \bm{u}}q\pb{\bm{u}}d\bm{u}}.
    \end{aligned}
\end{equation}
The approximate inducing value posterior, $q\pb{\bm{u}}$ is chosen as Cholesky variational distribution.

The multi-dimensional outputs are generated by multitasking variational strategy, with a constant mean for the location residual in both dimensions. The covariance matrix is computed based on the Matern kernel, with the smoothness parameter chosen as 5/2. The Adam optimizer is employed to maximize the marginal log likelihood
\begin{equation}
    \begin{aligned}
        p_{f}\pb{\bm{\delta}|\bm{U}} = \int{p\pb{\bm{\delta}| f\pb{\bm{U}}}p\pb{f\pb{\bm{U}}|\bm{U}}d\bm{u}}.
    \end{aligned}
\end{equation}
The model error bound is optimized by choosing variational evidence lower bound (ELBO) as loss function in SVGP.

\paragraph{Training procedures}
There are two ways to train the self-correction dynamic model.  If you already have an open-loop dynamic model, the RCM could be trained by utilizing the existing dynamic model to label the residual. Otherwise, the DM can be trained with the same raw data for RCM. One possible training process for the dynamic model is presented in \cite{apollo_2019}.

\subsection{DRF in simulation}
DRF is designed to fulfill the control-in-the-loop simulation, where a vehicle cannot always perfectly follow the instruction from planning due to dynamics and actuators' limitations. 
The simulation environment is available at \url{http://bce.apollo.auto/}.

\begin{figure}[h]
    \centering
    \includegraphics[width=0.48\textwidth]{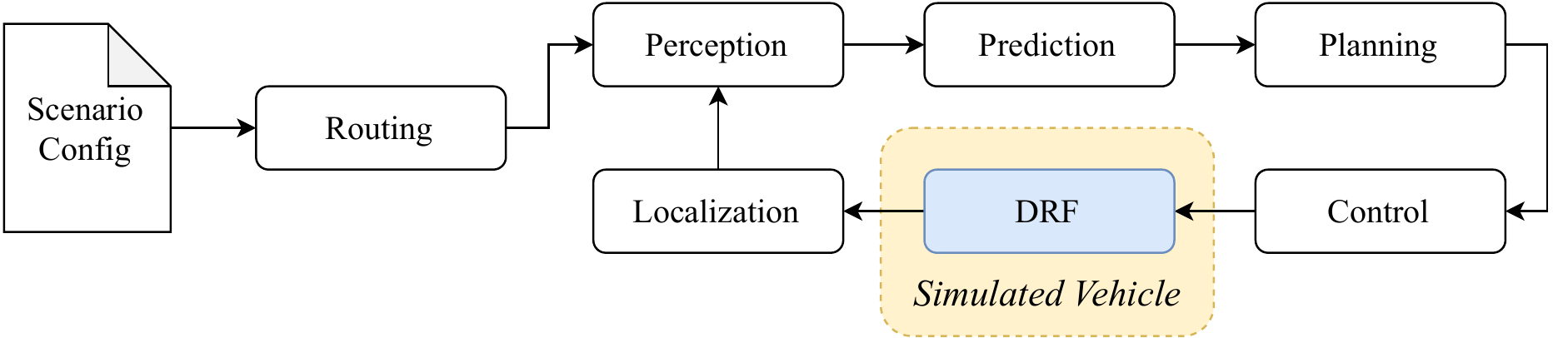}
    \caption{Control-in-the-Loop Simulation}
    \label{fig:simulation_schema}
\end{figure}

As shown in Fig. \ref{fig:simulation_schema}, in simulation, at each time cycle, there is a loop formed by perception, prediction, planning, control and localization modules. The control module takes inputs from upstream planning module, and generates commands to the simulated vehicle.  DRF updates states and the pose of the virtual vehicle based on control commands and vehicle states. The outputs of DRF are fed to localization module and utilized by simulation for the next time cycle. An accurate simulated vehicle should behave the same as a real vehicle when being fed with the same inputs. 
\begin{figure}[!htb]
    \centering
    \begin{minipage}{0.43\textwidth}
        \centering
        \captionsetup{justification=centering}
        \subfloat[Right turn\label{figure:right_turn}]{\includegraphics[width=.75\linewidth]{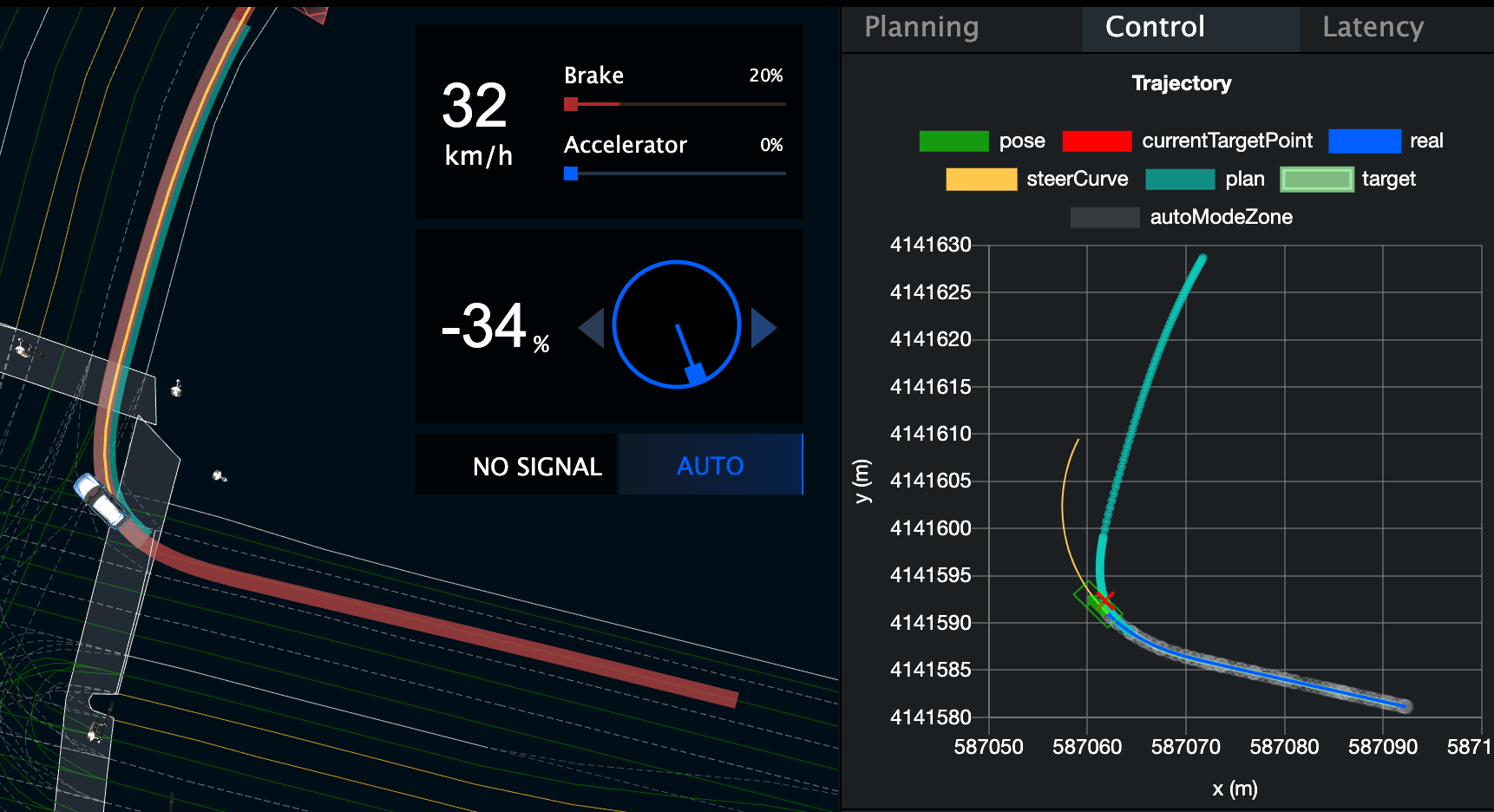}}
    \end{minipage}
    \quad
    \begin{minipage}{0.43\textwidth}
        \centering
        \captionsetup{justification=centering}
        \subfloat[U-turn\label{figure:u_turn}]{\includegraphics[width=.75\linewidth]{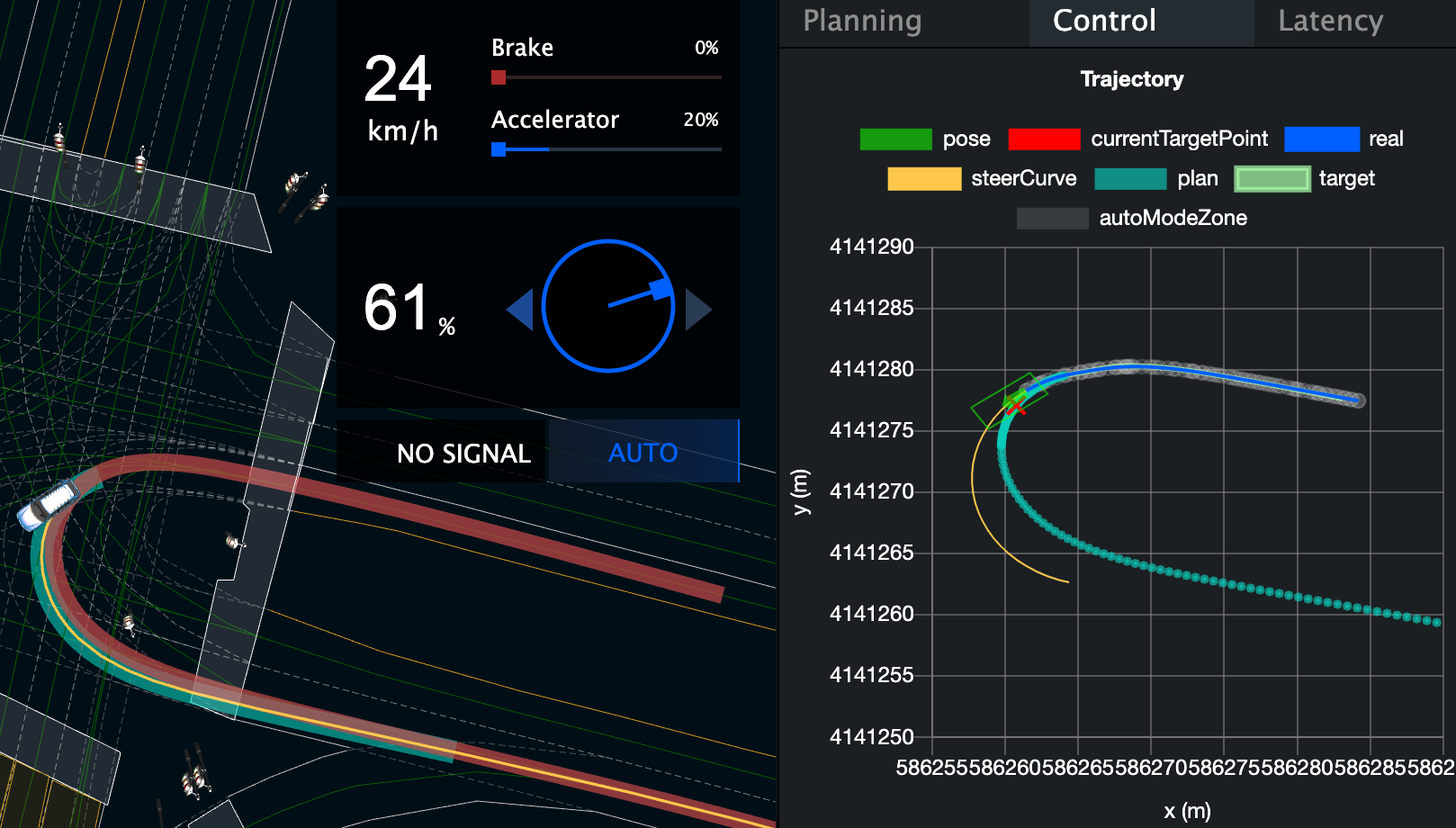}}
    \end{minipage}
    \caption{DRF in simulation}
    \label{fig:dfr_in_simulation}
\end{figure}

Fig. \ref{fig:dfr_in_simulation} shows DRF performing a control-in-the-loop right turn and U-turn respectively in simulation.  Just like the real vehicle, the simulated vehicle (the white car), as expected, does not perfectly follow the target trajectory from the planning module under current control commands.

\subsection{Performance Evaluation}
To remove feed-back control effects, we propose a non-feedback control-in-the-loop evaluation method, which injects control commands from road test to dynamic models. The accuracy of DRF is measured by the similarity between the trajectories from DRF and the ground truth under the same control commands.

\paragraph{Common Evaluation Metrics}
The cumulative absolute trajectory error, i.e., c-ATE, and mean absolute trajectory error i.e., m-ATE \cite{peretroukhin2017dpc} are defined as
\begin{equation}
\label{eq:cate}
    \begin{aligned}
        \epsilon_{\text{c-ATE}} = \sum_{i=1}^{N}\norm{\bm{p}_{m,i},\bm{p}_{gt,i}}_{2}, \\
         \epsilon_{\text{m-ATE}} = \frac{1}{N} \sum_{i=1}^{N}\norm{\bm{p}_{m,i},\bm{p}_{gt,i}}_{2},
    \end{aligned}
\end{equation}

where $\bm{p}_{gt, i}$ is the $i$th point of ground truth trajectory, i.e., vehicle trajectory from RTK (Real-time kinematic).  $\bm{p}_{m,i}$ is the $i$th point of model predicted trajectory. The $\norm{.}_{2}$ is used to measure the Euclidean distance between two points.   

\paragraph{Extra Evaluation Metrics}
Five extra evaluation metrics are chosen to further validate the similarity between DRF predicted trajectories and the ground truth trajectories.
The end-pose difference (ED) is the distance between the end locations of two trajectories. 

Two-sigma defect rate ($\epsilon_{2\sigma}$) is defined as the number of points with \emph{true} location residual (between ground truth location and DM predicted location, as in Fig. \ref{fig:DM20_structure}) falling out of the $2\sigma$ range of RCM \emph{predicted} location residual divided by the total number of trajectory points.
\begin{equation}
    \begin{aligned}
        \epsilon_{2\sigma} = 1-\frac{\sum_{i=1}^{N}\eta\pb{\bm{p}_{m,i}}}{N},
    \end{aligned}
\end{equation}
where $\eta\pb{\bm{p}_{m,i}}=1$ when $\bm{p}_{gt,i}\in \bm{p}_{m,i} + 2\sigma$ and $\eta\pb{\bm{p}_{m,i}}=0$ everywhere else. 

The Hausdirff Distance (HAU)~\cite{lee2007trajectory} between the model trajectory and the ground truth trajectory is
\begin{equation}
    \begin{aligned}
        \epsilon_{\text{HAU}}=\max\{\max_{\forall \bm{p}_{gt} \in T_{gt}} \min_{\forall \bm{p}_{m} \in T_{m}} \norm{\bm{p}_{gt}, \bm{p}_{m}}_{2}, \\
        \max_{\forall \bm{p}_{m} \in T_{m}} \min_{\forall \bm{p}_{gt} \in T_{gt}} \norm{\bm{p}_{gt}, \bm{p}_{m}}\}_{2}
    \end{aligned}
\end{equation}
where $\bm{p}_{m}$ is trajectory points from model predicted trajectory $T_{m}$ and $\bm{p}_{gt}$ is those of the ground truth, $T_{gt}$. 

The longest common sub-sequence error~\cite{buzan2004extraction} between the model output trajectory and the ground truth is
\begin{equation}
    \begin{aligned}
        \epsilon_{\text{LCSS}}=1 - \frac{L\pb{T_{m}-T_{gt}}}{\min{(N_{m},N_{gt})}}
    \end{aligned}
\end{equation}
where $L\pb{\cdot}$ is a function providing the common distance of the model trajectory and the ground truth trajectory. We choose the threshold in both $x$ direction and $y$ direction as $0.1$ meter.  $N_{m}$ and $N_{gt}$ are the length of the model trajectory and the ground truth trajectory respectively. Dynamic time warping (DTW)~\cite{keogh2000scaling}, a common method to measure the similarity between two temporal trajectories, is also chosen as a metric here.

\section{Experiments and Results}
\label{sec:experiment_and_results}
\subsection{Experiment Setup}
\label{sec:experiment_setup}
\begin{figure}[!htb]
    \centering
    \begin{minipage}{0.23\textwidth}
        \centering
        \captionsetup{justification=centering}
        \subfloat[MKZ with Apollo System\label{fig:Mkz7_pic}]{\includegraphics[width=.95\linewidth]{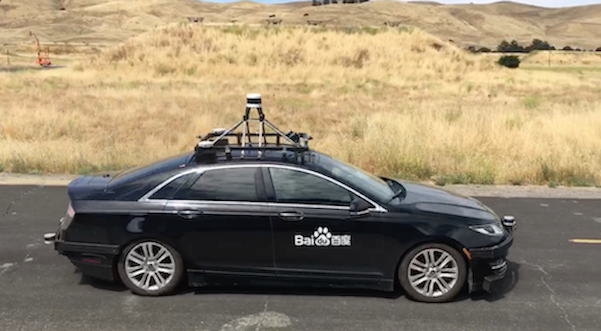}}
    \end{minipage}
    \begin{minipage}{0.23\textwidth}
        \centering
        \captionsetup{justification=centering}
        \subfloat[MKZ Parameters\label{fig:Mkz7_param}]{\includegraphics[width=.95\linewidth]{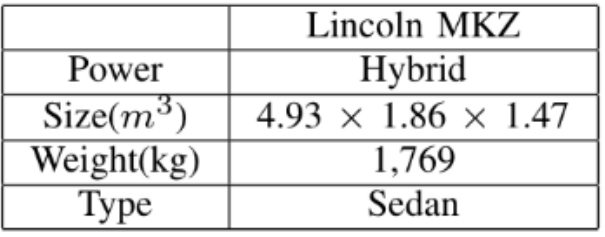}}
    \end{minipage}
    \caption{Vehicle in Experiments}
\end{figure}

We test DRF on a Lincoln MKZ model vehicle as shown in
Fig. \ref{fig:Mkz7_pic}, with its parameters listed in Fig. \ref{fig:Mkz7_param}.  The vehicle is equipped with Apollo autonomous driving system, which provides data collection GUI and cloud-based data logging system.  The training and evaluation datasets are collected by both manual and autonomous driving modes in urban roads.

Training data for DRF are augmented by adding an $85\%$ overlapping between adjacent data points. The augmented data are then categorized by its control command values and vehicle speed, as shown in Fig.~\ref{fig:collection_finish.png}, and re-sampled to achieve a uniform distribution over all categories.  The processed data include $330,012$ data points in total, which is about $91.67$-hour driving. The data is split into training/validation datasets for each category in the ratio of $8:2$.



\subsection{Performance Study}
\label{sec:performance_study}
The performance of seven models: two traditional dynamic models and five variations of DRF with structures shown in Table \ref{table:encoder_params}, are studied.  The performance metrics and evaluation scenarios are introduced in~\ref{sec:experiment_setup}. 

\begin{figure}[!h]
    \centering
    \begin{minipage}{0.23\textwidth}
        \centering
        \captionsetup{justification=centering}
        \subfloat[\label{fig:batch}]{\includegraphics[width=.95\linewidth]{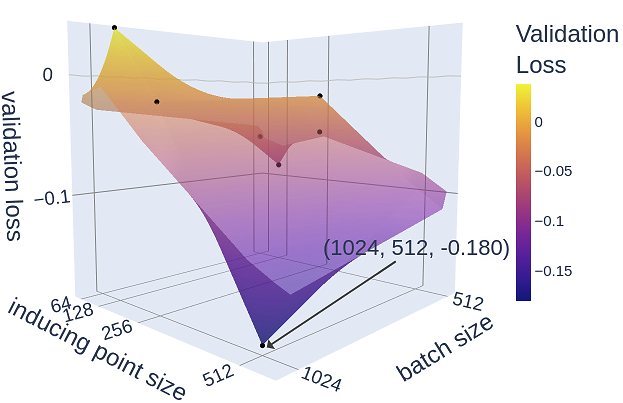}}
    \end{minipage}
    \begin{minipage}{0.23\textwidth}
        \centering
        \captionsetup{justification=centering}
        \subfloat[ \label{fig:lr}]{\includegraphics[width=.95\linewidth]{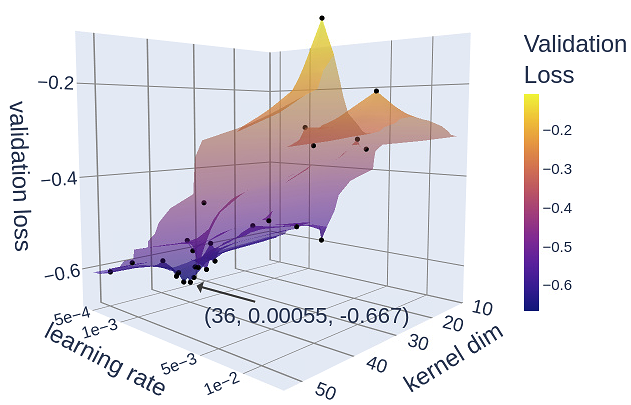}}
    \end{minipage}
    \begin{minipage}{0.23\textwidth}
        \centering
        \captionsetup{justification=centering}
        \subfloat[ \label{fig:head}]{\includegraphics[width=.95\linewidth]{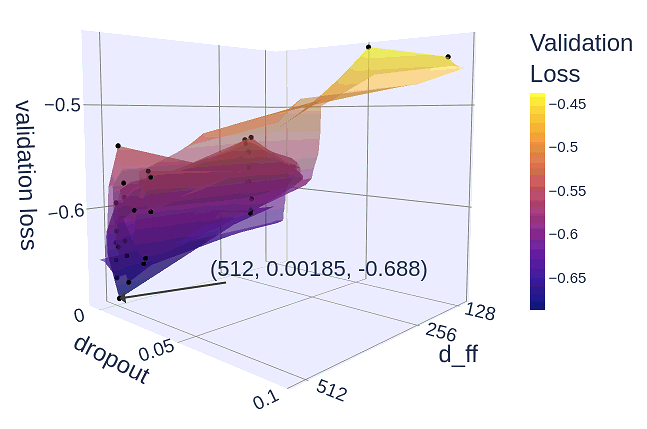}}
    \end{minipage}
        \begin{minipage}{0.23\textwidth}
        \centering
        \captionsetup{justification=centering}
        \subfloat[ \label{fig:validation}]{\includegraphics[width=.95\linewidth]{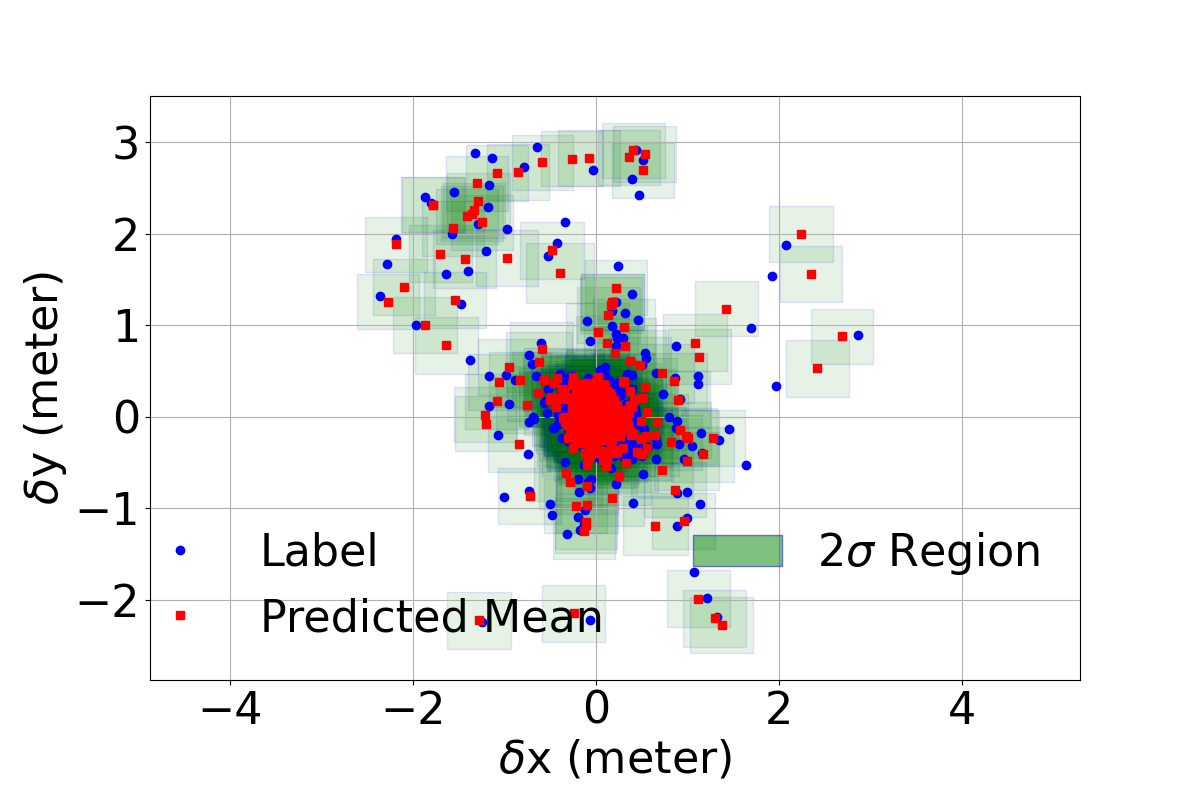}}
    \end{minipage}
    \caption{DRF-TRANS Ablation Test. Validation loss distribution with respect to (a) batch sizes and inducing point sizes (b) kernel dimensions and initial learning rates (c) dropout value and feed-forward layer dimensions. And (d) Model predictions on a validation dataset}
    \label{fig:autotune_res_tf}
\end{figure}

The hyperparameters of each DRF variation are tuned to achieve the minimum validation loss. Fig.~\ref{fig:autotune_res_tf} shows the validation loss of DRF-TRANS decreasing by tuning ~\emph{batch size},~\emph{inducing points},~\emph{feed-forward dimension},~\emph{dropout rate},~\emph{kernel size} and~\emph{initial learning rate}. 


\subsubsection{Performance across Scenarios}

\begin{figure}[!h]
    \centering
    \begin{minipage}{0.23\textwidth}
        \centering
        \captionsetup{justification=centering}
        \subfloat[Left Turn \label{fig:goldenset_left_turn}]{\includegraphics[width=.95\linewidth]{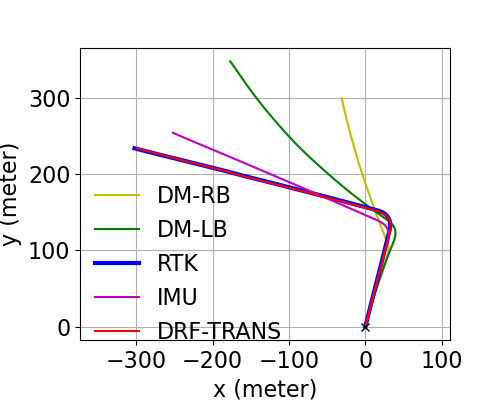}}
    \end{minipage}
    \begin{minipage}{0.23\textwidth}
        \centering
        \captionsetup{justification=centering}
        \subfloat[Left U-turn \label{fig:goldenset_left_u_turn}]{\includegraphics[width=.95\linewidth]{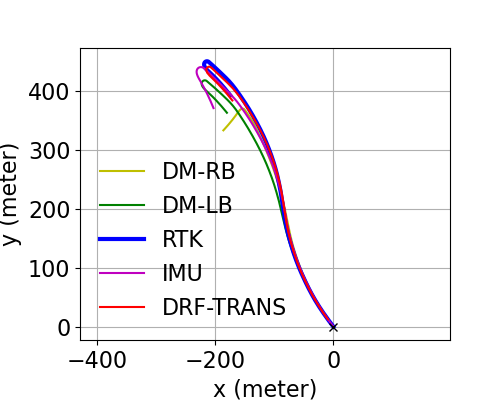}}
    \end{minipage}
    \begin{minipage}{0.23\textwidth}
        \centering
        \captionsetup{justification=centering}
        \subfloat[$2\sigma$ Region of Left turn \label{fig:goldenset_left_turn}]{\includegraphics[width=.95\linewidth]{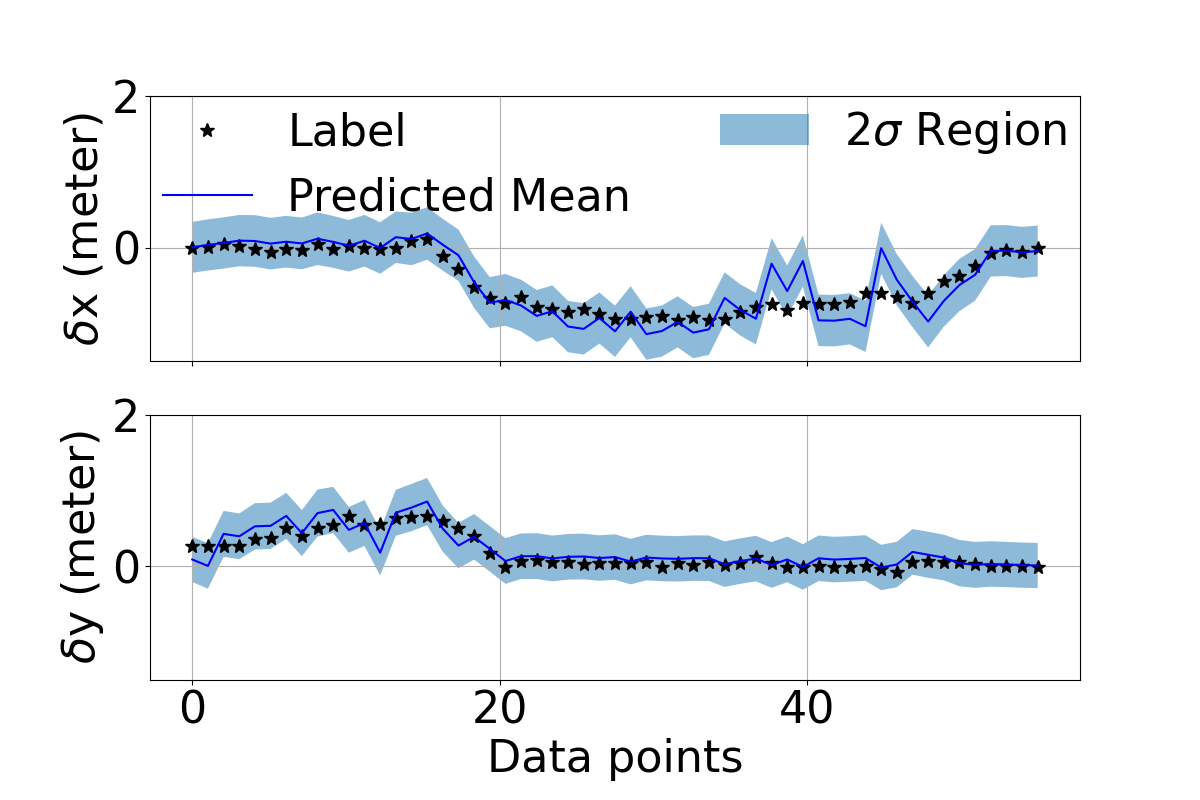}}
    \end{minipage}
    \begin{minipage}{0.23\textwidth}
        \centering
        \captionsetup{justification=centering}
        \subfloat[$2\sigma$ Region of Left U-turn \label{fig:goldenset_left_u_turn}]{\includegraphics[width=.95\linewidth]{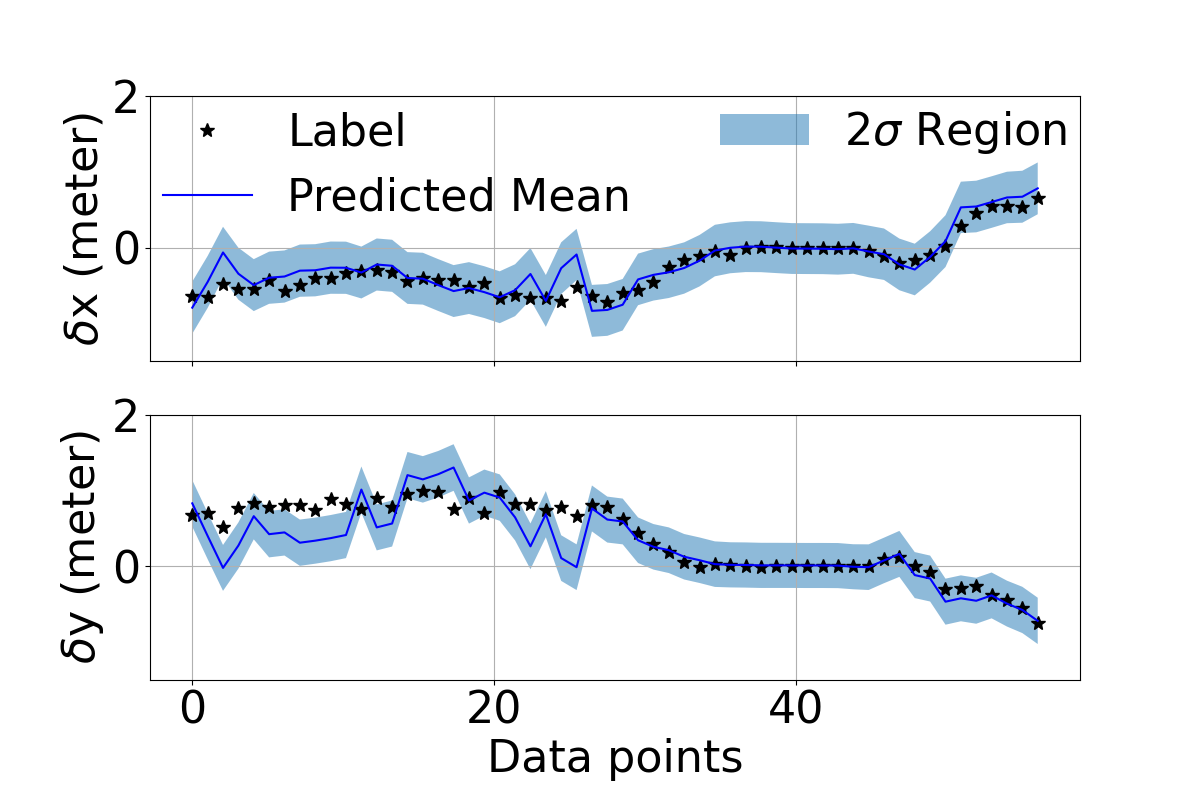}}
    \end{minipage}
    \caption{Evaluation results of DRF-TRANS in the most and least accuracy improved scenarios.}
    \label{fig:goldenset_best_worst}
\end{figure}

Table. \ref{table:cross_scenario_ate} is the c-ATE  and m-ATE comparisons among different models averaged over the golden evaluation set. Table \ref{table:per_scenario_ate} (in Appendix) is  m-ATE comparison for each scenario in the golden evaluation set. When averaged across scenarios, DRF-CNN has an error drop of $67.84\%$ to $76.00\%$ in c-ATE  and m-ATE at $1$s, $5$s, $10$s, $30$s and EoT (End of Trajectory) compared with DM-LB, a minimum drop of $29.37\%$ in left u-turn scenario EoT, and a maximum drop of $91.02\%$ in left turn scenario EoT. For the best performance model, DRF-TRANS, these numbers are $74.12\%$ to $85.02\%$ compared with DM-LB, $62.33\%$ in left u-turn scenario EoT and $97.52\%$ in left turn scenario EoT. Fig. \ref{fig:goldenset_best_worst} shows the detailed model performance comparison over the most and least improved scenarios, as well as the comparison of ground truth data vs $2\sigma$ range of predicted data.

\begin{table}[!ht]
    \caption{Performance averaged  across scenarios. DM-RB is rule-based dynamic model. DM-LB is open-loop learning-based dynamic model.  DRF-CNN, DRF-DCNN (Dilated CNN), DRF-ATTEN , DRF-LSTM and DRF-TRANS are variations of DRF.} 
    \centering
    \begin{tabular}{l|lllll}
        \hline
        Metrics    & \multicolumn{5}{c}{c-ATE (meter)}                                                                         \\ \hline
        Time  & 1  (sec)                               & 5 (sec)          & 10 (sec)              & 30 (sec)             & EoT
        \\ \hhline{=:=====}
        DM-RB         & 0.596                             & 14.185         & 79.885          & 1097.192        & 4301.295         \\
        DM-LB         & 0.510                             & 9.616          & 46.773          & 501.747         & 1880.532         \\
        DRF-CNN        & 0.164                             & 3.776          & 19.436          & 160.596         & 474.848          \\
        DRF-DCNN       & 0.231                             & 4.474          & 19.723          & 135.697         & 414.970          \\
        DRF-ATTEN      & 0.391                             & 6.186          & 24.113          & 200.227         & 613.305          \\
        DRF-LSTM       & 0.463                             & 7.005          & 25.862          & 187.784         & 571.349          \\
        DRF-TRANS      & \textbf{0.133}                    & \textbf{2.430} & \textbf{10.919} & \textbf{93.823} & \textbf{302.569} \\
        \hhline{=:=====}
        Metrics    & \multicolumn{5}{c}{m-ATE (meter)}                                                                         \\ \hline
        Time  & 1 (sec)                                 & 5 (sec)              & 10 (sec)              & 30 (sec)              & EoT
        \\ \hhline{=:=====}
        DM-RB         & 0.298                             & 2.364          & 7.262           & 35.393          & 73.583           \\
        DM-LB         & 0.255                             & 1.603          & 4.252           & 16.185          & 34.009           \\
        DRF-CNN        & 0.082                             & 0.629          & 1.767           & 5.181           & 8.161            \\
        DRF-DCNN       & 0.115                             & 0.746          & 1.793           & 4.377           & 7.125            \\
        DRF-ATTEN      & 0.195                             & 1.031          & 2.192           & 6.459           & 10.531           \\
        DRF-LSTM       & 0.231                             & 1.167          & 2.351           & 6.058           & 9.806            \\
        DRF-TRANS      & \textbf{0.066}                    & \textbf{0.405} & \textbf{0.993}  & \textbf{3.027}  & \textbf{5.093}   \\
        \hline
    \end{tabular}
    \label{table:cross_scenario_ate}
\end{table}


Besides c-ATE and m-ATE, we also show the performance of different models with extra evaluation metrics in Table \ref{table:other_metrics}.  We show that the best performed model over scenarios is DRF-TRANS with an error drop ratio of up to $88.48\%$ (in ED) compared with DM-LB. 


\begin{table}[!ht]
    \caption{Performance Evaluation under other metrics, EoT}
    \begin{tabular}{p{0.4cm}|p{0.6cm}p{1.6cm}p{0.4cm}p{0.7cm}p{0.7cm}}
        \hline
                 Model                  & $\epsilon_{\text{ED}}$             & $\pb{\epsilon_{2\sigma,x}, \epsilon_{2\sigma,y}}$            & $\epsilon_{\text{LCSS}}$            & $\epsilon_{\text{DTW}}$              & $\epsilon_{\text{HAU}}$              \\
                                                    & (meter)           &             &         & (meter)      & (meter)            \\
                 \hhline{=:=====}
        \multicolumn{1}{l|}{DM-RB}    & 140.330        & N/A         & 0.981          & 3793.037         & 149.731                \\
        \multicolumn{1}{l|}{DM-LB}    & 77.952         & N/A          & 0.981       & 1708.289         & 78.206                 \\
        \multicolumn{1}{l|}{DRF-CNN}   & 13.271         & (0.305, 0.036)          & 0.967         & 310.394          & 13.743                   \\
        \multicolumn{1}{l|}{DRF-DCNN}  & 12.651         & (0.237, 0.018)          &  0.966          & 269.528          & 12.823                  \\
        \multicolumn{1}{l|}{DRF-ATTEN} & 17.169         & (0.276, 0.064)           & 0.979        & 382.824          & 18.038                  \\
        \multicolumn{1}{l|}{DRF-LSTM}  & 15.835         & (0.251, 0.082)          & 0.978          & 353.148          & 16.876                 \\
        \multicolumn{1}{l|}{DRF-TRANS} & \textbf{8.981} & \textbf{(0.230, 0.009)} & \textbf{0.959} & \textbf{209.962} & \textbf{9.177}  \\ \hline
    \end{tabular}
    \label{table:other_metrics}
\end{table}


\subsubsection{Performance over Sunnyvale Loop}

\begin{figure}[!h]
    \centering
    \includegraphics[width=0.4\textwidth]{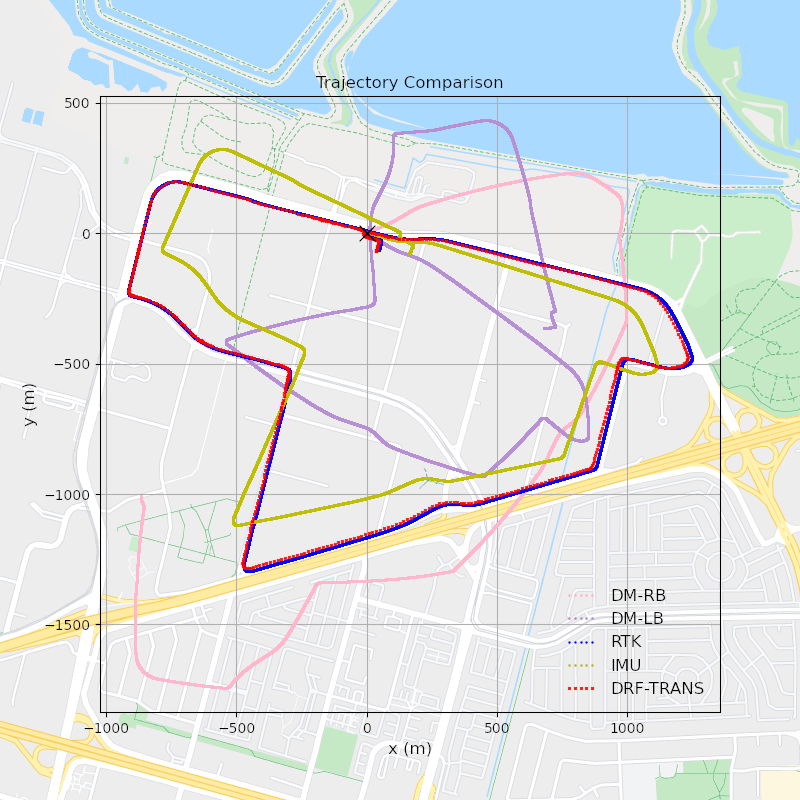}
    \caption[test]{Ground truth and model predicted trajectories over Sunnyvale Loop. The trajectory predicted by DRF-TRANS almost overlapped with the ground truth, and much more accurate than that of DM-RB or DM-LB}
    \label{fig:big_loop_1}
\end{figure}

Fig. \ref{fig:big_loop_1} is a 10-minute driving route. The vehicle starts from parking lot and drives on a regular road with speed variations at stop signs, traffic lights, and makes right/left turns. We use RTK trajectory as the ground truth and plot the integrated results for IMU, DM-LB, and our DRF-TRANS. We see the new model has a $98.53\%$ drop of both c-ATE and m-ATE at the end of this route compared with DM-LB.

\begin{figure}[!h]
    \centering
    \begin{minipage}{0.23\textwidth}
        \centering
        \captionsetup{justification=centering}
        \subfloat[m-ATE results\label{fig:loop_mate}]{\includegraphics[width=.95\linewidth]{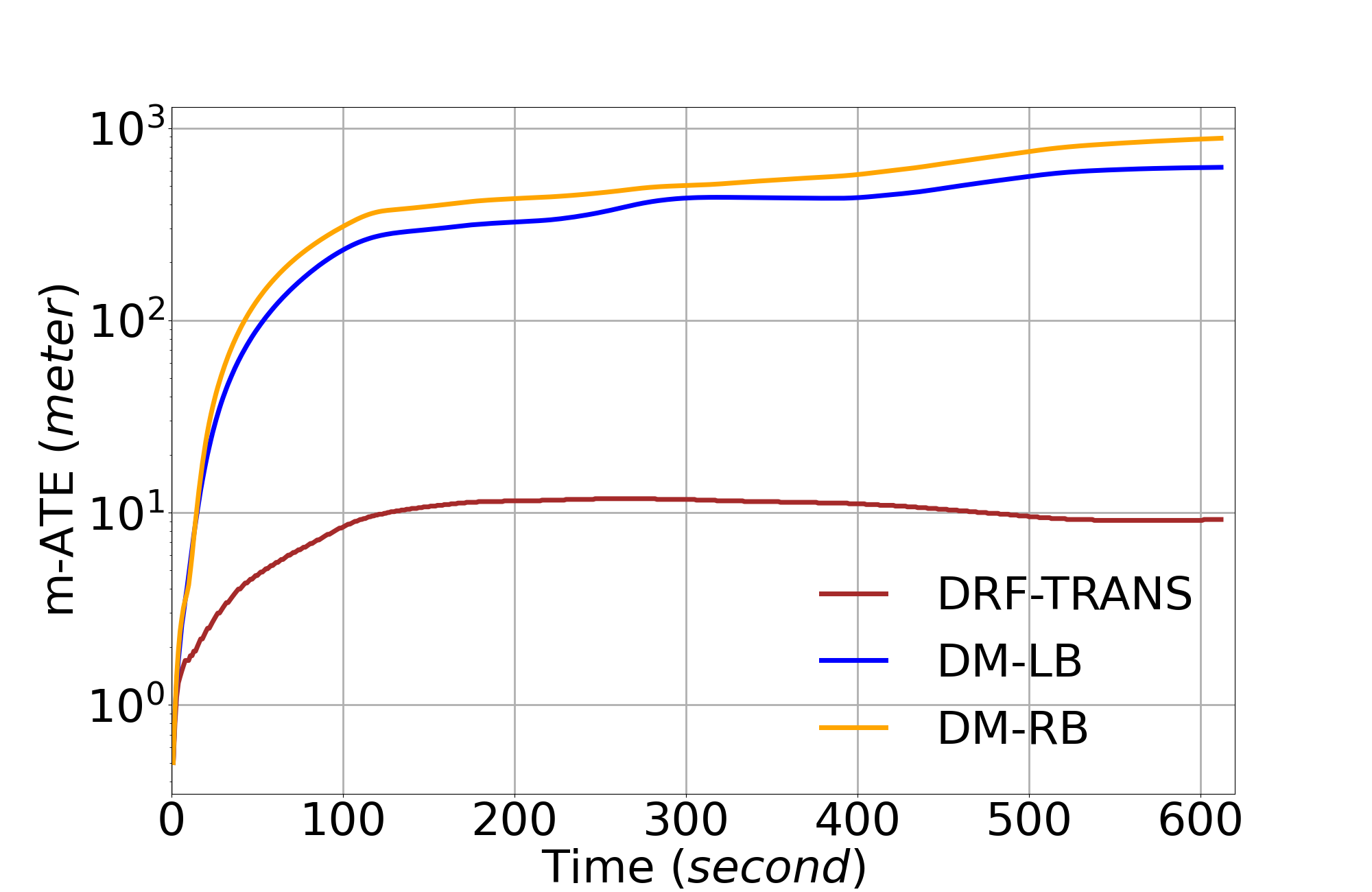}}
    \end{minipage}
    \begin{minipage}{0.23\textwidth}
        \centering
        \captionsetup{justification=centering}
        \subfloat[c-ATE results\label{fig:loop_cate}]{\includegraphics[width=.95\linewidth]{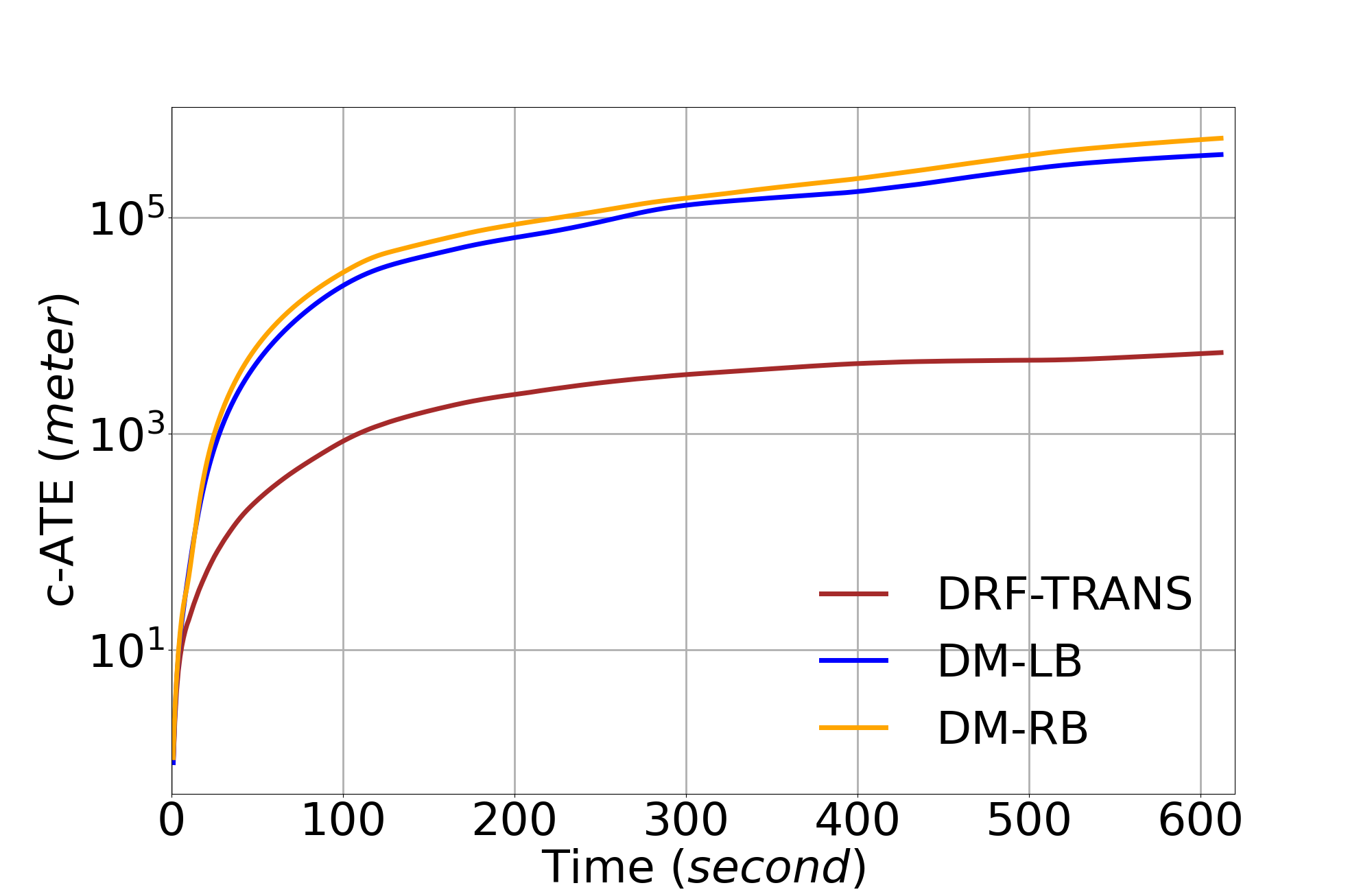}}
    \end{minipage}
    \caption{Performance comparison between DRF-TRANS, DM-RB and DM-LB over Sunnyvale Loop.}
    \label{fig:goldenset}
\end{figure}
\section{Conclusions}





In this paper, we presented a new learning-based vehicle dynamic
modeling framework (DRF) to bridge the the gap between simulation
and real world in autonomous driving. This framework consists of two
parts: 1) an open-loop dynamic model, either rule-based or learning-based, which takes control commands
and predicts vehicle future dynamics status; 2) a residual correction model composed of deep encoder network and SVGP, which works as a compensation
module to overcome the prediction residual from the open-loop dynamic
model. With changeable deep encoder networks, five vehicle dynamic models are derived from DRF. We show that the best performed model among these five achieves 
an average $5.093m$ m-ATE in evaluation scenarios, with a drop of $85.02\%$ compared
with using open-loop dynamic model only. We also show our models outperform the open-loop one across five additional evaluation metrics. DRF also
provides error bounds of location prediction, which could be utilized as important reference information when testing planning and control algorithms. All these features give DRF a high potential for widespread application from related industries to academia. In the future, this design could be further improved by extending residual correction to the heading angle and other vehicle dynamic states. 





\begin{appendices}
    \section{}

\begin{table}[!h]
    \caption{m-ATE (meter) on scenarios:1) left turn; 2) left turn with stop; 3) right turn; 4) right turn with stop; 5) left U-turn; 6) right U-turn; 7) zig-zag starting from left; 8) zig-zag starting from right.}
    \label{table:per_scenario_ate}
    \begin{tabular}{ll|lllll}
        \hline
                      \multicolumn{2}{c|}{ }   & 1 (sec)  & 5 (sec)    & 10 (sec)    & 30 (sec)    & EoT     
                                              \\ \hhline{==:=====}
        \multicolumn{1}{c}{\multirow{7}{*}{1}} & DM-RB    & 0.120 & 1.027 & 4.463  & 30.837 & 108.822 \\
        \multicolumn{1}{c}{}                   & DM-LB   & 0.127 & 0.773 & 2.383  & 13.876 & 60.120  \\
        \multicolumn{1}{c}{}                   & DRF-CNN   & 0.118 & 0.403 & 0.847  & 2.933  & 5.396   \\
        \multicolumn{1}{c}{}                   & DRF-DCNN  & 0.113 & 0.558 & 0.925  & 2.263  & 4.824   \\
        \multicolumn{1}{c}{}                   & DRF-ATTEN & 0.098 & 0.799 & 1.986  & 5.059  & 7.488   \\
        \multicolumn{1}{c}{}                   & DRF-LSTM  & 0.100 & 0.801 & 1.977  & 6.004  & 9.226   \\
        \multicolumn{1}{c}{}                   & DRF-TRANS & \textbf{0.102} & \textbf{0.268} & \textbf{0.248}  & \textbf{0.862}  & \textbf{1.489}   \\ \hline
        \multirow{7}{*}{2}                     & DM-RB    & 0.233 & 1.982 & 6.637  & 36.541 & 57.517  \\
                                              & DM-LB   & 0.216 & 1.392 & 4.426  & 26.602 & 41.252  \\
                                              & DRF-CNN   & 0.035 & 0.246 & 0.606  & 4.353  & 5.922   \\
                                              & DRF-DCNN  & 0.032 & 0.119 & 0.397  & 2.314  & 3.230   \\
                                              & DRF-ATTEN & 0.162 & 0.611 & 1.279  & 3.266  & 5.071   \\
                                              & DRF-LSTM  & 0.243 & 0.853 & 1.632  & 4.291  & 6.796   \\
                                              & DRF-TRANS & \textbf{0.010} & \textbf{0.089} & \textbf{0.456}  & \textbf{2.001}  & \textbf{2.844}   \\ \hline
        \multirow{7}{*}{3}                     & DM-RB    & 0.505 & 3.270 & 8.324  & 36.540 & 54.261  \\
                                              & DM-LB   & 0.449 & 2.530 & 6.048  & 16.704 & 19.846  \\
                                              & DRF-CNN   & 0.029 & 1.126 & 3.513  & 6.834  & 7.969   \\
                                              & DRF-DCNN  & 0.056 & 1.092 & 2.951  & 5.464  & 6.478   \\
                                              & DRF-ATTEN & 0.345 & 1.780 & 3.679  & 9.324  & 11.992  \\
                                              & DRF-LSTM  & 0.409 & 2.217 & 4.717  & 10.119 & 12.370  \\
                                              & DRF-TRANS & \textbf{0.062} & \textbf{1.188} & \textbf{2.500}  & \textbf{4.574}  & \textbf{5.410}   \\ \hline
        \multirow{7}{*}{4}                     & DM-RB    & 0.496 & 3.667 & 10.013 & 45.067 & 71.425  \\
                                              & DM-LB   & 0.424 & 2.720 & 7.201  & 23.169 & 51.103  \\
                                              & DRF-CNN   & 0.033 & 0.728 & 3.480  & 8.344  & 11.866  \\
                                              & DRF-DCNN  & 0.075 & 0.886 & 3.468  & 8.959  & 13.062  \\
                                              & DRF-ATTEN & 0.258 & 1.909 & 4.768  & 12.525 & 17.449  \\
                                              & DRF-LSTM  & 0.330 & 2.184 & 5.078  & 12.073 & 16.651  \\
                                              & DRF-TRANS & \textbf{0.063} & \textbf{0.294} & \textbf{1.780}  & \textbf{4.759}  & \textbf{6.657}   \\ \hline
        \multirow{7}{*}{5}                     & DM-RB    & 0.565 & 3.611 & 8.922  & 36.706 & 62.649  \\
                                              & DM-LB   & 0.497 & 2.641 & 5.843  & 20.775 & 27.367  \\
                                              & DRF-CNN   & 0.104 & 1.089 & 3.527  & 14.144 & 19.339  \\
                                              & DRF-DCNN  & 0.322 & 2.031 & 4.544  & 11.944 & 15.505  \\
                                              & DRF-ATTEN & 0.365 & 1.775 & 3.456  & 10.485 & 15.145  \\
                                              & DRF-LSTM  & 0.420 & 1.940 & 3.206  & 6.281  & 8.895   \\
                                              & DRF-TRANS & \textbf{0.025} & \textbf{0.488} & \textbf{1.465}  & \textbf{7.118}  & \textbf{10.307}  \\ \hline
        \multirow{7}{*}{6}                     & DM-RB    & 0.289 & 1.899 & 5.041  & 24.848 & 43.816  \\
                                              & DM-LB   & 0.261 & 1.526 & 3.702  & 8.480  & 13.848  \\
                                              & DRF-CNN   & 0.240 & 1.000 & 1.254  & 1.430  & 4.005   \\
                                              & DRF-DCNN  & 0.277 & 1.016 & 1.281  & 1.029  & 3.501   \\
                                              & DRF-ATTEN & 0.277 & 0.849 & 0.957  & 4.444  & 9.099   \\
                                              & DRF-LSTM  & 0.266 & 0.786 & 0.874  & 4.409  & 8.813   \\
                                              & DRF-TRANS & \textbf{0.211} & \textbf{0.602} & \textbf{0.678}  & \textbf{0.412}  & \textbf{0.775}   \\ \hline
        \multirow{7}{*}{7}                     & DM-RB    & 0.017 & 1.276 & 7.027  & 43.865 & 87.116  \\
                                              & DM-LB   & 0.010 & 0.593 & 2.388  & 14.658 & 33.173  \\
                                              & DRF-CNN   & 0.046 & 0.207 & 0.287  & 1.926  & 4.268   \\
                                              & DRF-DCNN  & 0.014 & 0.067 & 0.203  & 1.653  & 4.351   \\
                                              & DRF-ATTEN & 0.035 & 0.234 & 0.623  & 4.113  & 8.666   \\
                                              & DRF-LSTM  & 0.004 & 0.153 & 0.413  & 2.863  & 6.698   \\
                                              & DRF-TRANS & \textbf{0.030} & \textbf{0.155} & \textbf{0.239}  & \textbf{2.847}  & \textbf{6.289}   \\ \hline
        \multirow{7}{*}{8}                     & DM-RB    & 0.159 & 2.182 & 7.670  & 28.742 & 103.058 \\
                                              & DM-LB   & 0.054 & 0.648 & 2.026  & 5.218  & 25.364  \\
                                              & DRF-CNN   & 0.050 & 0.235 & 0.620  & 1.480  & 6.523   \\
                                              & DRF-DCNN  & 0.034 & 0.197 & 0.575  & 1.392  & 6.052   \\
                                              & DRF-ATTEN & 0.024 & 0.291 & 0.788  & 2.455  & 9.342   \\
                                              & DRF-LSTM  & 0.079 & 0.407 & 0.911  & 2.420  & 8.999   \\
                                              & DRF-TRANS & \textbf{0.029} & \textbf{0.155} & \textbf{0.574}  & \textbf{1.639}  & \textbf{6.976}   \\ \hline  

    \end{tabular}
\end{table}

\end{appendices}

\clearpage
\bibliographystyle{IEEEtran}

\bibliography{IEEEabrv,./refs}

\begin{thebibliography}{10}
\providecommand{\url}[1]{#1}
\csname url@samestyle\endcsname
\providecommand{\newblock}{\relax}
\providecommand{\bibinfo}[2]{#2}
\providecommand{\BIBentrySTDinterwordspacing}{\spaceskip=0pt\relax}
\providecommand{\BIBentryALTinterwordstretchfactor}{4}
\providecommand{\BIBentryALTinterwordspacing}{\spaceskip=\fontdimen2\font plus
\BIBentryALTinterwordstretchfactor\fontdimen3\font minus
  \fontdimen4\font\relax}
\providecommand{\BIBforeignlanguage}[2]{{%
\expandafter\ifx\csname l@#1\endcsname\relax
\typeout{** WARNING: IEEEtran.bst: No hyphenation pattern has been}%
\typeout{** loaded for the language `#1'. Using the pattern for}%
\typeout{** the default language instead.}%
\else
\language=\csname l@#1\endcsname
\fi
#2}}
\providecommand{\BIBdecl}{\relax}
\BIBdecl

\bibitem{gevers2006linearID}
M.~Gevers, ``A personal view of the development of system identification: A
  30-year journey through an exciting field,'' in \emph{IEEE Control systems
  magazine}, vol.~26, no.~6, 2006, pp. 93--105.

\bibitem{gordon1993novel}
N.~J. Gordon, D.~J. Salmond, and A.~F. Smith, ``Novel approach to
  nonlinear/non-gaussian bayesian state estimation,'' in \emph{IEEE proceedings
  F (radar and signal processing)}, vol. 140, no.~2.\hskip 1em plus 0.5em minus
  0.4em\relax IET, 1993, pp. 107--113.

\bibitem{nelles2013nonlinear}
O.~Nelles, \emph{Nonlinear system identification: from classical approaches to
  neural networks and fuzzy models}.\hskip 1em plus 0.5em minus 0.4em\relax
  Springer Science \& Business Media, 2013.

\bibitem{rong2006sequential}
H.-J. Rong, N.~Sundararajan, G.-B. Huang, and P.~Saratchandran, ``Sequential
  adaptive fuzzy inference system (safis) for nonlinear system identification
  and prediction,'' \emph{Fuzzy sets and systems}, vol. 157, no.~9, pp.
  1260--1275, 2006.

\bibitem{ghahramani1999learning}
Z.~Ghahramani and S.~T. Roweis, ``Learning nonlinear dynamical systems using an
  em algorithm,'' in \emph{Advances in neural information processing systems},
  1999, pp. 431--437.

\bibitem{Agudelo2020dataset}
D.~Agudelo-Espana, A.~Zadaianchuk, P.~Wenk, A.~Garg, J.~Akpo, F.~Grimminger,
  J.~Viereck, M.~Naveau, L.~Righetti, G.~Martius \emph{et~al.}, ``A real-robot
  dataset for assessing transferability of learned dynamics models,'' in
  \emph{2020 IEEE International Conference on Robotics and Automation
  (ICRA)}.\hskip 1em plus 0.5em minus 0.4em\relax IEEE, 2020, pp. 8151--8157.

\bibitem{Deisenroth201pilco}
M.~Deisenroth and C.~E. Rasmussen, ``Pilco: A model-based and data-efficient
  approach to policy search,'' in \emph{Proceedings of the 28th International
  Conference on machine learning (ICML-11)}, 2011, pp. 465--472.

\bibitem{romeres2019semiparam}
D.~Romeres, D.~K. Jha, A.~DallaLibera, B.~Yerazunis, and D.~Nikovski,
  ``Semiparametrical gaussian processes learning of forward dynamical models
  for navigating in a circular maze,'' in \emph{2019 International Conference
  on Robotics and Automation (ICRA)}.\hskip 1em plus 0.5em minus 0.4em\relax
  IEEE, 2019, pp. 3195--3202.

\bibitem{hensman2015scalable}
J.~Hensman, A.~Matthews, and Z.~Ghahramani, ``Scalable variational gaussian
  process classification,'' \emph{In International Conference on Artificial
  Intelligence and Statistics}, 2015.

\bibitem{brossard2019learning}
M.~Brossard and S.~Bonnabel, ``Learning wheel odometry and imu errors for
  localization,'' in \emph{2019 International Conference on Robotics and
  Automation (ICRA)}, 2019, pp. 291--297.

\bibitem{su2020deepnn}
H.~Su, W.~Qi, C.~Yang, J.~Sandoval, G.~Ferrigno, and E.~De~Momi, ``Deep neural
  network approach in robot tool dynamics identification for bilateral
  teleoperation,'' \emph{IEEE Robotics and Automation Letters}, vol.~5, no.~2,
  pp. 2943--2949, 2020.

\bibitem{zhang2020sufficient}
C.~Zhang, A.~Khan, S.~Paternain, and A.~Ribeiro, ``Sufficiently accurate model
  learning,'' in \emph{2020 IEEE International Conference on Robotics and
  Automation (ICRA)}.\hskip 1em plus 0.5em minus 0.4em\relax IEEE, 2020, pp.
  10\,991--10\,997.

\bibitem{Williams2017MPC}
G.~Williams, N.~Wagener, B.~Goldfain, P.~Drews, J.~M. Rehg, B.~Boots, and E.~A.
  Theodorou, ``Information theoretic mpc for model-based reinforcement
  learning,'' in \emph{2017 IEEE International Conference on Robotics and
  Automation (ICRA)}.\hskip 1em plus 0.5em minus 0.4em\relax IEEE, 2017, pp.
  1714--1721.

\bibitem{Punjani2015helicopter}
A.~Punjani and P.~Abbeel, ``Deep learning helicopter dynamics models,'' in
  \emph{2015 IEEE International Conference on Robotics and Automation (ICRA)},
  2015, pp. 3223--3230.

\bibitem{georgiou2015predicting}
T.~Georgiou and Y.~Demiris, ``Predicting car states through learned models of
  vehicle dynamics and user behaviours,'' in \emph{2015 IEEE Intelligent
  Vehicles Symposium (IV)}, 2015, pp. 1240--1245.

\bibitem{Devineau2018vehicle}
G.~Devineau, P.~Polack, F.~Altch{\'e}, and F.~Moutarde, ``Coupled longitudinal
  and lateral control of a vehicle using deep learning,'' in \emph{2018 21st
  International Conference on Intelligent Transportation Systems (ITSC)}.\hskip
  1em plus 0.5em minus 0.4em\relax IEEE, 2018, pp. 642--649.

\bibitem{apollo_2019}
J.~{Xu}, Q.~{Luo}, K.~{Xu}, X.~{Xiao}, S.~{Yu}, J.~{Hu}, J.~{Miao}, and
  J.~{Wang}, ``An automated learning-based procedure for large-scale vehicle
  dynamics modeling on baidu apollo platform,'' in \emph{2019 IEEE/RSJ
  International Conference on Intelligent Robots and Systems (IROS)}, 2019, pp.
  5049--5056.

\bibitem{chang2017dilated}
S.~Chang, Y.~Zhang, W.~Han, M.~Yu, X.~Guo, W.~Tan, X.~Cui, M.~Witbrock, M.~A.
  Hasegawa-Johnson, and T.~S. Huang, ``Dilated recurrent neural networks,'' in
  \emph{Advances in Neural Information Processing Systems}, 2017, pp. 77--87.

\bibitem{parmar2019stand}
N.~Parmar, P.~Ramachandran, A.~Vaswani, I.~Bello, A.~Levskaya, and J.~Shlens,
  ``Stand-alone self-attention in vision models,'' in \emph{Advances in Neural
  Information Processing Systems}, 2019, pp. 68--80.

\bibitem{sun20183dof}
L.~Sun, Z.~Yan, S.~M. Mellado, M.~Hanheide, and T.~Duckett, ``3dof pedestrian
  trajectory prediction learned from long-term autonomous mobile robot
  deployment data,'' in \emph{2018 IEEE International Conference on Robotics
  and Automation (ICRA)}.\hskip 1em plus 0.5em minus 0.4em\relax IEEE, 2018,
  pp. 1--7.

\bibitem{vaswani2017attention}
A.~Vaswani, N.~Shazeer, N.~Parmar, J.~Uszkoreit, L.~Jones, A.~N. Gomez,
  {\L}.~Kaiser, and I.~Polosukhin, ``Attention is all you need,'' in
  \emph{Advances in neural information processing systems}, 2017, pp.
  5998--6008.

\bibitem{gardner2018gpytorch}
J.~R. Gardner, G.~Pleiss, D.~Bindel, K.~Q. Weinberger, and A.~G. Wilson,
  ``Gpytorch: Blackbox matrix-matrix gaussian process inference with gpu
  acceleration,'' in \emph{Advances in Neural Information Processing Systems},
  2018, pp. 7576--7586.

\bibitem{peretroukhin2017dpc}
V.~Peretroukhin and J.~Kelly, ``Dpc-net: Deep pose correction for visual
  localization,'' \emph{IEEE Robotics and Automation Letters}, vol.~3, no.~3,
  pp. 2424--2431, 2017.

\bibitem{lee2007trajectory}
J.-G. Lee, J.~Han, and K.-Y. Whang, ``Trajectory clustering: a
  partition-and-group framework,'' in \emph{Proceedings of the 2007 ACM SIGMOD
  international conference on Management of data}, 2007, pp. 593--604.

\bibitem{buzan2004extraction}
D.~Buzan, S.~Sclaroff, and G.~Kollios, ``Extraction and clustering of motion
  trajectories in video,'' in \emph{Proceedings of the 17th International
  Conference on Pattern Recognition, 2004. ICPR 2004.}, vol.~2.\hskip 1em plus
  0.5em minus 0.4em\relax IEEE, 2004, pp. 521--524.

\bibitem{keogh2000scaling}
E.~J. Keogh and M.~J. Pazzani, ``Scaling up dynamic time warping for datamining
  applications,'' in \emph{Proceedings of the sixth ACM SIGKDD international
  conference on Knowledge discovery and data mining}, 2000, pp. 285--289.

\end{thebibliography}

\end{document}